\pdfoutput=1
\documentclass[11pt]{article}

\usepackage[]{EMNLP2023}

\usepackage{times}
\usepackage{latexsym}

\usepackage{booktabs}

\usepackage[T1]{fontenc}
\usepackage[utf8]{inputenc}

\usepackage{microtype}

\usepackage{inconsolata}
\usepackage{pgfplots}
\usepackage{pgf}
\usepackage{subcaption}
\usepackage{amsmath}
\usepackage{booktabs}
\usepackage{multirow}
\usepackage{algorithm}
\usepackage{algpseudocode}
\usepackage{adjustbox}
\usepackage{graphicx}
\usepackage{comment}
\usepackage{subcaption}
\usepackage{hyperref}

\pgfplotsset{every axis/.append style={
                    xlabel={$x$},          % default put x on x-axis
                    ylabel={$y$},          % default put y on y-axis
                    label style={font=\sffamily\small},
                    tick label style={font=\sffamily\small},
                    xticklabel style = {font=\sffamily\scriptsize},
                    yticklabel style = {font=\sffamily\scriptsize},
                    title style = {font=\sffamily\footnotesize},
                    ylabel near ticks,
                    y label style={font=\sffamily\scriptsize},
                    xlabel near ticks,
                    x label style={font=\sffamily\scriptsize},
                    legend cell align={left},
                    legend style={draw=none, font=\sffamily\scriptsize},
                    },
                    legend image code/.code={
                    \draw[mark repeat=2,mark phase=2]
                        plot coordinates {
                        (0cm,0cm)
                        (0.15cm,0cm)        %% default is (0.3cm,0cm)
                        (0.3cm,0cm)         %% default is (0.6cm,0cm)
                        };%
                    }
                    }
\pgfplotsset{compat=newest}

\definecolor{darkgray176}{RGB}{176,176,176}
\definecolor{darkorange25512714}{RGB}{255,127,14}
\definecolor{forestgreen4416044}{RGB}{44,160,44}
\definecolor{lightgray204}{RGB}{204,204,204}
\definecolor{steelblue}{RGB}{31,119,180}

\newcommand{\method}{TK-KNN }

\title{{\textbf{TK-KNN: A Balanced Distance-Based Pseudo Labeling Approach for Semi-Supervised Intent Classification}}}
\author{Nicholas Botzer$^\clubsuit$, David Vasquez$^\spadesuit$, Tim Weninger$^\clubsuit$, Issam Laradji$^\spadesuit$ \\
  $^\clubsuit$University of Notre Dame, $^\spadesuit$ServiceNow Research \\
  \texttt{$^\clubsuit$nbotzer@nd.edu, $^\spadesuit$issam.laradji@gmail.com}
}

\date{}

\begin{document}

\maketitle

\begin{abstract}
The ability to detect intent in dialogue systems has become increasingly important in modern technology.
These systems often generate a large amount of unlabeled data, and manually labeling this data requires substantial human effort.
Semi-supervised methods attempt to remedy this cost by using a model trained on a few labeled examples and then by assigning pseudo-labels to further a subset of unlabeled examples that has a model prediction confidence higher than a certain threshold. However, one particularly perilous consequence of these methods is the risk of picking an imbalanced set of examples across classes, which could lead to poor labels. In the present work, we describe Top-K K-Nearest Neighbor (TK-KNN), which uses a more robust pseudo-labeling approach based on distance in the embedding space while maintaining a balanced set of pseudo-labeled examples across classes through a ranking-based approach.
Experiments on several datasets show that TK-KNN outperforms existing models, particularly when labeled data is scarce on popular datasets such as CLINC150 and Banking77. Code is available at \url{https://github.com/ServiceNow/tk-knn}
\end{abstract}

% (Maximum 3400 characters - about 1.5 pages)
\section{Introduction.}
% 1. What is the goal?
Large language models like BERT~\citep{devlin2018bert} have significantly pushed the boundaries of Natural Language Understanding (NLU) and created interesting applications such as automatic ticket resolution~\citep{marcuzzo2022multi}. A key component of such systems is a virtual agent's ability to understand a user's intent to respond appropriately. Successful implementation and deployment of models for these systems require a large amount of labeled data to be effective. Although deployment of these systems often generate a large amount of data that could be used for fine-tuning, the cost of labeling this data is high.
%However, finetuning them on new datasets first require collecting a decent amount of labels which translates to high human effort.
Semi-supervised learning methodologies are an obvious solution because they can significantly reduce the amount of human effort required to train these kinds of models~\citep{laradji2021ssr,zhu2009introduction} especially in image classification tasks~\citep{zhai2019s4l, ouali2020semi, yalniz2019billion}. However, as well shall see, applications of these models is difficult for NLU and intent classification because of the label distribution.
% 2. Where are we now?

\begin{figure}
    \centering
    \includegraphics[width=\linewidth]{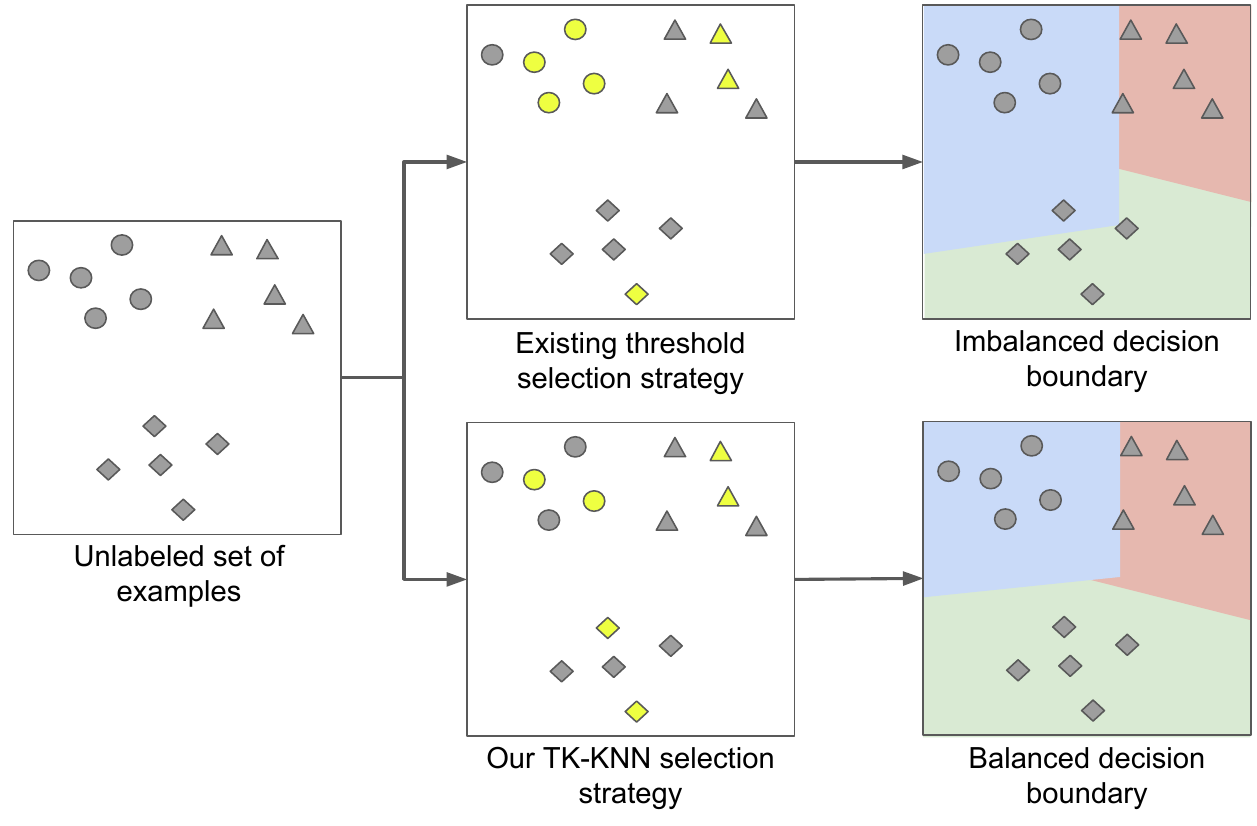}
    \caption{Example of pseudo label selection when using a threshold (top) versus the top-k sampling strategy (bottom). In this toy scenario, we chose $k=2$, where each class is represented by a unique shape. As the threshold selection strategy pseudo-labels data elements (shown as yellow) that exceed the confidence level, the model tends to become biased towards classes that are easier to predict. This bias causes a cascade of mis-labels that leads to even more bias towards the majority class.}
    \label{fig:teaser}
\end{figure}

Indeed, research most closely realted to the present work is the Slot-List model by \citet{basu2021semi}, which focuses on the meta-learning aspect of semi-supervised learning rather than using unlabeled data.
In a similar vein the GAN-BERT~\citep{croce2020gan} model  shows that using an adversarial learning regime can be devised to ensure that the extracted BERT features are similar amongst the unlabeled and the labeled data sets and substantially boost classification performance.
Other methods have investigated how data augmentation can be applied to the NLP domain to enforce consistency in the models~\citep{chen2020mixtext}, and several other methods have been proposed from the computer vision community. However, a recent empirical study found that many of these methods do not provide the same benefit to NLU tasks as they provide to computer vision tasks~\citep{chen2021empirical} and can even hinder performance.% in certain instances.

% 3. Why is it hard?
Intent classification remains a challenging problem for multiple reasons. Generally, the number of intents a system must consider is relatively large, with sixty classes or more. On top of that, most queries consists of only a short sentence or two. This forces models to need many examples in order to learn nuance between different intents within the same domain. 
In the semi-supervised setting, many methods set a confidence threshold for the model and assign pseudo-labels to the unlabeled data if their confidence is above the threshold~\citep{sohn2020fixmatch}. This strategy permits high-confidence pseudo-labeled data elements to be included in the training set, which typically results in performance gains.
Unfortunately, this approach also causes the model to become overconfident for classes that are easier to predict.
%The issue is more pronounced for intent classification because of feedback loops that can quickly cause the model to become biased towards a small number of classes.

% 4. What do we do to solve it?
In the present work, we describe the Top-K K-Nearest Neighbor (TK-KNN) method for training semi-supervised models. The main idea of this method is illustrated in Figure~\ref{fig:teaser}. TK-KNN makes two improvements over other pseudo-labeling approaches. First, to address the model overconfidence problem, we use a top-k sampling strategy when assigning pseudo-labels to enforce a balanced set of classes by taking the top-k predictions per class, not simply the predictions that exceed a confidence threshold overall predictions. Furthermore, when selecting the top-k examples the sampling strategy does not simply rely on the model's predictions, which tend to be noisy. Instead we leverage the embedding space of the labeled and unlabeled examples to find those with similar embeddings and combine them with the models' predictions.
Experiments using standard performance metrics of intent classification are performed on three datasets: CLINC150~\citep{larson2019evaluation}, Banking77~\citep{casanueva2020efficient}, and Hwu64~\citep{liu2019benchmarking}. We find that the TK-KNN method outperforms existing methods in all scenarios and performs exceptionally well in the low-data scenarios.

% In this work we look into a similar challenge. We are given a training dataset consisting of labelled artificially created or out-of-distribution  chat logs and incidents, and the goal is to leverage unlabeled data so that the model performs better intent classification on the test set. There is a large distribution shift between the training set and the test set, but the unlabeled set has a similar to the distribution as the test set. Compared to   GAN-BERT, we look into comparing different methods that fall under contrastive learning~\cite{jaiswal2021survey} and test them on a set of standard datasets like CLINC150, SNIPS, and ATIS. Contributing to this area opens doors for new possibilities in minimizing human effort and using transfer learning to address the task of intent classification. This task is of great interest to many industries and the research community that work on resolving dialogues or ticketing systems that require understanding of customer's intent.

\section{Related Work}

\paragraph{Intent Classification}
The task of intent classification has attracted much attention in recent years due to the increasing use of virtual customer service agents. Recent research into intent classification systems has mainly focused on learning out of distribution data~\citep{zhan2021out, zhang2021deep, cheng2022learning, zhou2022knn}. These techniques configure their experiments to learn from a reduced number of the classes and treat the remaining classes as out-of-distribution during testing. Although this research is indeed important in its own regard, it deviates from the present work's focus on semi-supervised learning.

\paragraph{Pseudo Labeling}
Pseudo labeling is a mainstay in semi-supervised learning~\citep{lee2013pseudo, rizve2021defense, cascante2021curriculum}. In simple terms, pseudo labelling uses the model itself to acquire hard labels for each of the unlabeled data elements. This is achieved by taking the argmax of the models' output and treating the resulting label as the example's label. In this learning regime, the hard labels are assigned to the unlabeled examples without considering the confidence of the model's predictions. These pseudo-labeled examples are then combined with the labeled data to train the model iteratively. The model is then expected to iteratively improve until convergence. The main drawback of this method is that mislabeled data elements early in training can severely degrade the performance of the system.

A common practice to help alleviate mislabeled samples is to use a threshold $\tau$ to ensure that only high-quality (\textit{i.e.}, confident) labels are retained~\citep{sohn2020fixmatch}. 
The addition of confidence restrictions into the training process~\citep{sohn2020fixmatch} has shown improvements but also restricts the data used at inference time and introduces the confidence threshold value as yet another hyperparameter that needs to be tuned.

Another major drawback of this selection method is that the model can become very biased towards the easy classes in the early iterations of learning~\citep{arazo2020pseudo}.
Recent methods, such as FlexMatch~\citep{zhang2021flexmatch}, have discussed this problem at length and attempted to address this issue with a curriculum learning paradigm that allows each class to have its own threshold. These thresholds tend to be higher for majority classes lower for less-common classes. However, this only serves to exacerbate the problem because the less-common classes will have less-confident labels. A previous work by ~\citet{zou2018unsupervised} proposes a similar class balancing parameter to be learned per class, but is applied to the task of unsupervised domain adaptation.
A close previous work to ours is co-training \cite{nigam2000analyzing} that iteratively adds a single example from each class throughout the self-training. Another more recent work ~\citep{gera2022zero} also proposes a balanced sampling mechanism for self-training, but starts from a zero-shot perspective and limits to two cycles of self-training.

Another pertinent work is by ~\citet{chen2022contrastnet}, who introduce ContrastNet, a framework that leverages contrastive learning for few-shot text classification. This is particularly relevant to our study considering the challenges posed by datasets with a scarce number of labeled examples per class. A notable work by ~\citet{wang2022contrastive} employs a contrastive learning-enhanced nearest neighbor mechanism for multi-label text classification, which bears some resemblance to the KNN strategies discussed in our work. 

The TK-KNN strategy described in the present work addresses these issues by learning the decision boundaries for all classes in a balanced way while still giving preference to accurate labels by considering the proximity between the labeled and the unlabeled examples in the embedding space.

\paragraph{Distance-based Pseudo labeling}
Another direction explored in recent work is to consider the smoothness and clustering assumptions found in semi-supervised learning~\citep{ouali2020overview} for pseudo labeling. The smoothness assumption states that if two points lie in a high-density region their outputs should be the same. The clustering assumption similarly states that if points are in the same cluster, they are likely from the same class.
Recent work by \citet{zhu2022detecting} propose a training-free approach to detect corrupted labels. They use a k-style approach to detect corrupted labels that share similar features. The results of this work show that the smoothness and clustering assumptions are also applicable in a latent embedding space and therefore data elements that are close in the latent space are likely to share the same clean label.

Two other recent works have made use of these assumptions in semi-supervised learning to improve their pseudo-labeling process.
First, \citet{taherkhani2021self} use the Wasserstein distance to match clusters of unlabeled examples to labeled clusters for pseudo-labeling.

Second, the aptly-named feature affinity based pseudo-labeling~\citep{ding2019feature} method uses the cosine similarity between unlabeled examples and cluster centers that have been discovered for each class. The selected pseudo label is determined based on the highest similarity score calculated for the unlabeled example.

Results from both of these works demonstrate that distance-based pseudo-labeling strategies yield significant improvements over previous methods. However, both of these methods depend on clusters formed from the labeled data. In the intent classification task considered in the current study, the datasets sometimes have an extremely limited number of labeled examples per class, with instances where there is only one labeled example per class. This scarcity of labeled data makes forming reliable clusters quite challenging. Therefore, the TK-KNN model described in the present work adapted the K-Nearest Neighbors search strategy to help guide our pseudo-labeling process.

\section{Top-K KNN Semi-Supervised Learning} 

%\begin{comment}
\begin{figure*}
    \includegraphics[width=\linewidth]{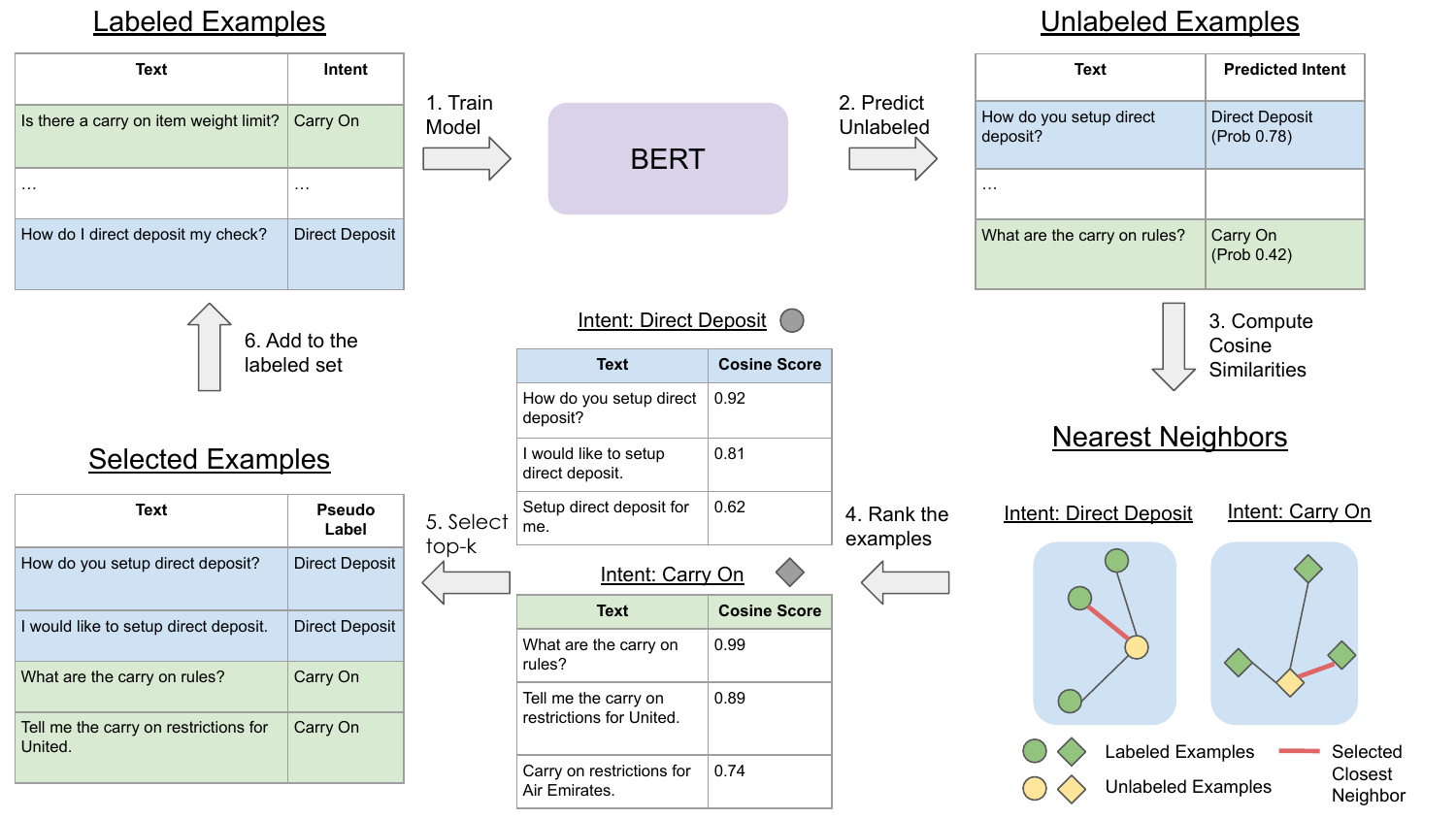}    
    \centering
    \caption{TK-KNN overview. The model is (1) trained on the small portion of labeled data. Then, this model is used to predict (2) pseudo labels on the unlabeled data. Then the cosine similarity (3) is calculated for each unlabeled data point with respect to the labeled data points in each class. Yellow shapes represent unlabeled data and green represent labeled data. Similarities are computed and unlabeled examples are ranked (4) based on a combination of their predicted probabilities and cosine similarities. Then, the top-k ($k=2$) examples are selected (5) for each class. These examples are finally added (6) to the labeled dataset to continue the iterative learning process.}
    \label{fig:self_training}
\end{figure*}
%\end{comment}

\subsection{Problem Definition}
We formulate the problem of semi-supervised intent classification as follows:

Given a set of labeled intents $X = {(x_n, y_n)}: n \in (1,...,N)$ where $x_n$ represents the intent example and $y_n$ the corresponding intent class $c \in C$ and a set of unlabeled intents $U = {u_m}: m \in (1,...,M)$, where each instance $u_m$ is an intent example lacking a label. Intents are fed to the model as input and the model outputs a predicted intent class, denoted as $p_{model}(c|x, \theta)$, where $\theta$ represents some pre-trained model parameters. Our goal is to learn the optimal parameters for $\theta$.

\subsection{Method Overview}
As described above, we first employ pseudo labeling to iteratively train (and re-train) a model based on its most confident-past predictions. In the first training cycle, the model is trained on only the small portion of labeled data $X$. In the subsequent cycles, the model is trained on the union of $X$ and a subset of the unlabeled data $U$ that has been pseudo-labeled by the model in the previous cycle. Figure ~\ref{fig:self_training} illustrates an example of this training regime with the TK-KNN method.

We use the BERT-base~\cite{devlin2018bert} model with an added classification head to the top. The classification head consists of a dropout layer followed by a linear layer with dropout and ends with an output layer that represents the dataset's class set $C$. We select the BERT-base model for fair comparison with other methods.

\subsection{Top-K Sampling}
When applying pseudo-labeling, it is often observed that some classes are easier to predict than others. In practice, this causes the model to become biased towards the easier classes~\citep{arazo2020pseudo} and perform poorly on the more difficult ones. The Top-K sampling process within the TK-KNN system seeks to alleviate this issue by growing the pseudo-label set across all labels together.

%To do this, we implement a \textbf{top-k} strategy for selecting the pseudo labels at each cycle of pseudo-labeling. 
%We train the model to convergence with the small initial labeled data in our implementation. 
When we perform pseudo labeling, we select the \textbf{top-k} predictions per class from the unlabeled data. 
This selection neither uses nor requires any threshold; instead, it limits each class to choose the predictions with the highest confidence.
We rank each predicted data element with a score based on the models predicted probability.
%We then assign each predicted data element a score based on the model's predicted probability.

\begin{equation}
    \textrm{score}(u_m) = p_{model}(y=c | u_m;\theta)
\end{equation}

%After each training cycle, we will have grown the pseudo labels in the dataset by $k$ times the number of classes at a time. 
After each training cycle, the number of pseudo labels in the dataset will have increased by $k$ times the number of classes.
This process continues until all examples are labeled or some number of pre-defined cycles has been reached. We employ standard early stopping criteria~\citep{prechelt1998early} during each training cycle to determine whether or not to stop training.

\subsection{KNN-Alignment}
Although our top-k selection strategy helps alleviate the model's bias, it still relies entirely on the model predictions. To enhance our top-k selection strategy, we utilize a KNN search to modify the scoring function that is used to rank which pseudo-labeled examples should be included in the next training iteration. The intuition for the use of the KNN search comes from the findings in~\cite{zhu2022detecting} where "closer" instances are more likely to share the same label based on the neighborhood information when some labels are corrupted, which often occurs in semi-supervised learning from the pseudo-labeling strategy.

%Modified version
Specifically, we extract a latent representation from each example in our training dataset, both the labeled and unlabeled examples. We formulate this latent representation in the same way as Sentence-BERT \cite{reimers2019sentence} to construct a robust sentence representation. This representation is defined as the mean-pooled representation of the final BERT layer that we formally define as:

\begin{equation}
\label{eq:mean-pooled}
    z = \texttt{mean}([CLS], T_1, T_2, ..., T_M)
\end{equation}

Where CLS is the class token, $T$ is each token in the sequence, $M$ is the sequence length, and $z$ is the extracted latent representation. When we perform our pseudo labeling process we extract the latent representation for all of our labeled data $X$ as well as our unlabeled data $U$.

%For each unlabeled example we calculate the cosine similarity of it's latent representation $l_m$ with the latent representation of each of each of labeled counterparts $l_n$ that belong to the predicted class. The cosine similarity score between the two representations is then extracted from this comparison as follows.

%For each unlabeled example we calculate the cosine similarity of it's latent representation with the latent representation of each of labeled counterparts that belong to the predicted class.

For each unlabeled example, we calculate the cosine similarity between its latent representation and the latent representations of the labeled counterparts belonging to the predicted class.

%Specifically, we extract a latent representation $l_n$ from BERT from each labeled data element $x_n$ after the model has finished a training cycle. During pseudo labeling, we also extract the latent representation, $l_m$, of each unlabeled data element $u_m$ also from BERT. The unlabeled representations are then compared to each set of labeled data elements belonging to the same class. The cosine similarity score between the two representations is then extracted from this comparison.

% \begin{equation}
    % \textrm{cos}\left(l_n, l_m\right) = \frac { l_n \cdot  l_m}{|| l_n|| \cdot || l_m||}
% \end{equation}

%The best scoring cosine similarity between the unlabeled example and it's labeled neighbors is used for calculating the score of an unlabeled example.
The highest cosine similarity score between the unlabeled example and its labeled neighbors is used to calculate the score of an unlabeled example. Let $z_m$ and $z_n$ be the latent representations of the unlabeled data point $u_m$ and a labeled data point $x_n$, respectively.
An additional hyperparameter, $\beta$, permits the weighing of the model's prediction and the cosine similarity for the final scoring function.

\begin{equation} \label{eq:2}
\begin{aligned}
%\resizebox{0.85\linewidth}{!}{
    \textrm{score}(u_m) ={} & (1 - \beta)\times p_{model}(y | u_m;\theta)  + \\ & \beta\times\textrm{sim}(z_n, z_m)
%}
\end{aligned}
\end{equation}

In this equation $\textrm{sim}(z_n, z_m)$ computes the cosine similarity between the latent representations of $u_m$ and its closest labeled counterpart in the predicted class.
With these scores we then follow the previously discussed top-k selection strategy to ensure balanced classes. The addition of the K-nearest neighbor search helps us to select more accurate labels early in the learning process. We provide pseudo code for our pseudo-labeling strategy in Appendix Algorithm~\ref{alg:cap}. 
%After training BERT across all cycles, we simply use it on the test set for inference.

\subsection{Loss Function}
As we use the cosine similarity to help our ranking method we want to ensure that similar examples are grouped together in the latent space.While the cross entropy loss is an ideal choice for classification, as it incentivizes the model to produce accurate predictions, it does not guarantee that discriminative features will be learned \citep{elsayed2018large}, which our pseudo labeling relies on.
To address this issue, we supplemented the cross-entropy loss with a supervised contrastive loss \cite{khosla2020supervised} and a differential entropy regularization loss \cite{sablayrolles2019spreading}, and trained the model using all three losses jointly.

%To augment this, we added a supervised contrastive loss \cite{khosla2020supervised} and a differential entropy regularization loss \cite{sablayrolles2019spreading} and jointly train all three.

%This equation is just a generic cross-entropy  loss function
\begin{equation}
L_{CE} = -\sum_{i=1}^{C} y_i \log(\hat{y}_i)
\end{equation}

We select the supervised contrastive loss as shown in \cite{khosla2020supervised}. This ensures that our model with learn good discriminative features in the latent space that separate examples belonging to different classes. The supervised contrastive loss relies on augmentations of the original examples. To get this augmentation we simply apply a dropout layer of $0.2$  to the representations that we extract from the model. As is standard for the supervised contrastive loss we add a separate projection layer to our model to align the representations. The representations fed to the projection layer is the mean-pooled BERT representation as shown in Eq.~\ref{eq:mean-pooled}.
%This ensures that our model will learn good sentence representations that will be used to select similar examples.

\begin{equation}
   \small L_{SCL} = \sum_{i \in I} \frac{-1}{|P(i)|} \sum_{p \in P(i)} log \frac{sim(z_i,z_p) / \tau}{ \sum_{a \in A(i)} sim(z_i, z_a) / \tau}
\end{equation}

When adopting the contrastive loss previous works \cite{el2021training} have discussed how the model can collapse in dimensions as a result of the loss. We follow this work in adopting a differential entropy regularizer in order to spread the representations our more uniformly. The method we use is based on the \citet{kozachenko1987sample} differential entropy estimator:

\begin{equation}
L_{KoLeo} = -\frac{1}{N}\sum_{i=1}^{N} \log(p_i)
\end{equation}

Where $p_i = min_{(i \neq j)} || f(x_i) - f(x_j)||$. This regularization helps to maximize the distance between each point and its neighbors. By doing so it helps to alleviate the rank collapse issue. We combine this term with the cross-entropy and contrastive objectives, weighting it using a coefficient $\gamma$.

\begin{equation}
    L_{ALL} = L_{CE} + L_{SCL} + \gamma L_{KoLeo}
\end{equation}

The joint training of these individual components leads our model to have better discriminative features that are more robust, that results in improved generalization and prediction accuracy.

\section{Experiments}

\subsection{Experimental Settings}

\paragraph{Datasets}
We use three well-known benchmark datasets to test and compare the TK-KNN model against other models on the intent classification task. 
Our intent classification datasets are \textbf{CLINC150}~\citep{larson2019evaluation} that contains 150 in-domain intents classes from ten different domains and one out-of-domain class.
\textbf{BANKING77}~\citep{casanueva2020efficient} that contains 77 intents, all related to the banking domain. \textbf{HWU64}~\citep{liu2019benchmarking} which includes 64 intents coming from 21 different domains. Banking77 and Hwu64 do not provide validation sets, so we created our own from the original training sets. All datasets are in English.
A breakdown of each dataset is shown in Table~\ref{tab:dataset}.

\begin{table}[h]
\centering
\scriptsize{
%\resizebox{\linewidth}{!}{
\begin{tabular}{llllll}
\toprule
\textbf{Dataset}   & \textbf{Intents} & \textbf{Domain} & \textbf{Train} & \textbf{Val} & \textbf{Test} \\
\midrule
CLINC150  & 151     & 10     & 15,250   &   3,100    & 5,550          \\
Banking77 & 77      & 1      & 9,002   &   1,001   & 3,080          \\
Hwu64     & 64      & 21     & 8,884    &   1,076   & 1,076         \\
\bottomrule
\end{tabular}
}

\caption{Breakdown of the intent classification datasets. Note that BANKING77 and HWU64 do not provide validation sets, so we generated a validation set from the original training set.}
\label{tab:dataset}
\end{table}

We conducted our experiments with varying amounts of labeled data for each dataset. All methods are run with five random seeds and the mean average accuracy of their results are reported along with their 95\% confidence intervals~\cite{dror2018hitchhiker}. 
%This methodology permits tests of statistical significance. Reported results are therefore accompanied by 95\% confidence intervals.

\subsection{Baselines}
To perform a proper and thorough comparison of TK-KNN with existing methods, we implemented and repeated the experiments on the following models and strategies.

\begin{itemize}
    \item \textbf{Supervised}: Use only labeled portion of dataset to train the model without any semi-supervised training. This model constitutes a competitive lower bound of performance because of the limits in the amount of labeled data.
    \item \textbf{Pseudo Labeling (PL)} \cite{lee2013pseudo}: This strategy trains the model to convergence then makes predictions on all of the unlabeled data examples. These examples are then combined with the labeled data and used to re-train the model in an iterative manner.
    \item \textbf{Pseudo Labeling with Threshold (PL-T)} \cite{sohn2020fixmatch}: This process follows the pseudo labeling strategy but only selects unlabeled data elements which are predicted above a threshold $\tau$. We use a $\tau$ of 0.95 based on the findings from previous work. 
    \item \textbf{Pseudo Labeling with Flexmatch (PL-Flex)} \cite{zhang2021flexmatch}: Rather than using a static threshold across all classes, a dynamic threshold is used for each class based on a curriculum learning framework.
    \item \textbf{GAN-BERT} \cite{croce2020gan}: This method applies generative adversarial networks \cite{goodfellow2020generative} to a pre-trained BERT model. The generator is an MLP that takes in a noise vector. The output head added to the BERT model acts as the discriminator and includes an extra class for predicting whether a given data element is real or not.
    \item \textbf{MixText} \cite{chen2020mixtext}: This method extends the MixUp \cite{zhang2017mixup} framework to NLP and uses the hidden representation of BERT to mix together. The method also takes advantage of consistency regularization in the form of back translated examples.
    \item \textbf{\method}: The method described in the present work using top-k sampling with a weighted selection based on model predictions and cosine similarity to the labeled samples.
    %\item \textbf{Top-k Upper}: Top-k sampling method, but always select the correct pseudo-label. This model serves as an upper bound.
\end{itemize}

\subsection{Implementation Details}
Each method uses the BERT base model with a classification head attached. We use the base BERT implementation provided by Huggingface~\citealp{wolf2019huggingface}, that contains a total of 110M parameters. All models are trained for 30 cycles of self-training.
The models are optimized with the AdamW optimizer with a learning rate of 5e-5.
Each model is trained until convergence by early stopping applied according to the validation set. We use a batch size of 256 across experiments and limit the sequence length to 64 tokens. For TK-KNN, we set $k=6$, $\beta=0.75$, and $\gamma=0.1$ and report the results for these settings. An ablation study of these two hyperparameters is presented later. For details on the settings used for MixText please see Appendix~\ref{app:mix-text}.

\paragraph{Computational Use.}
In total we estimate that we used around 18,000 GPU hours for this project.  For the final experiments and ablation studies we estimate that the TK-KNN model used ~4400 GPU hours.
Experiments were carried out on Nvidia Tesla P100 GPUs that each had 12GB of memory and 16GB of memory.

\section{Results}

\setlength{\tabcolsep}{3pt}
\begin{table}
    \centering
    \scriptsize{
    %\resizebox{\linewidth}{!}{
    
    \begin{tabular}{l| llll }
    \toprule
        \multirow{2}{*} &  \multicolumn{4}{c}{\textbf{Percent Labeled}} \\
        \textbf{Method}  & \multicolumn{1}{c}{1\%} & \multicolumn{1}{c}{2\%} & \multicolumn{1}{c}{5\%} & \multicolumn{1}{c}{10\%}  \\ 
        
        \midrule
        \multirow{2}{*} & \multicolumn{4}{c}{\underline{CLINC150}} \\
        \textbf{Supervised} &  27.35 \tiny{$\pm 1.71$} & 49.15 \tiny$\pm 1.99$ & 67.96 \tiny$\pm 0.85$ & 75.05 \tiny$\pm 1.57$  \\
        \textbf{PL} &   24.51 \tiny$\pm 3.92$ & 48.58 \tiny$\pm 1.79$ & 69.19 \tiny$\pm 0.54$ & 76.92 \tiny$\pm 1.05$  \\
        \textbf{PL-T} &  39.05 \tiny$\pm 3.26$ & 56.65 \tiny$\pm 1.53$ & 71.25 \tiny$\pm 0.5$ & 79.29 \tiny$\pm 1.62$  \\
        \textbf{PL-Flex} &  42.81 \tiny$\pm 4.39$ & 60.07 \tiny$\pm 1.42$ & 73.42 \tiny$\pm 1.62$ & 78.86 \tiny$\pm 1.01$ \\
        \textbf{GAN-BERT} & 18.18 \tiny$\pm 0.0$ & 23.29 \tiny$\pm 11.42$ & 44.89 \tiny$\pm 24.39$ & 63.02 \tiny$\pm 25.1$  \\
        %\textbf{MixText} & 28.51 \tiny$\pm 9.29$ & 63.48 \tiny$\pm 0.76$ & \textbf{79.64} \tiny$\pm 0.77$ & \textbf{81.20} \tiny$\pm 1.36$ \\
        \textbf{MixText} & 12.86 \tiny$\pm 6.39$ & 37.93 \tiny$\pm 16.8$ & 61.39 \tiny$\pm 0.77$ & 74.29 \tiny$\pm 0.37$ \\
        %\textbf{\method(Ours Old)} & 49.76 \tiny$\pm 3.06$ & 63.96 \tiny$\pm 1.42$ & 74.11 \tiny$\pm 1.22$ & 79.55 \tiny$\pm 0.49$ \\
        \textbf{\method} & \textbf{53.73} \tiny$\pm 1.72$ & \textbf{65.87} \tiny$\pm 1.18$ & \textbf{74.31} \tiny$\pm 0.96$ & \textbf{79.45} \tiny$\pm 1.01$ \\
        
        \midrule
        \multirow{2}{*} & \multicolumn{4}{c}{\underline{BANKING77}} \\      
        \textbf{Supervised} &  34.73 \tiny$\pm 1.5$ & 47.51 \tiny$\pm 2.89$ & 70.27 \tiny$\pm 1.08$ & 80.82 \tiny$\pm 0.41$  \\
        \textbf{PL} &   29.09 \tiny$\pm 3.83$ & 45.16 \tiny$\pm 2.71$ & 69.69 \tiny$\pm 2.16$ & 80.26 \tiny$\pm 0.49$  \\
        \textbf{PL-T} &  35.12 \tiny$\pm 3.86$ & 51.67 \tiny$\pm 3.14$ & 71.16 \tiny$\pm 1.98$ & 81.88 \tiny$\pm 0.43$ \\
        \textbf{PL-Flex} & 40.04 \tiny$\pm 3.4$ & 54.18 \tiny$\pm 3.31$ & 73.43 \tiny$\pm 1.55$ & 82.54 \tiny$\pm 0.84$ \\
        \textbf{GAN-BERT} & 5.4 \tiny$\pm 9.16$ & 16.98 \tiny$\pm 21.73$ & 54.09 \tiny$\pm 29.56$ & 79.64 \tiny$\pm 1.39$ \\
        %\textbf{MixText} & 44.13 \tiny$\pm 2.65$ & \textbf{76.94} \tiny$\pm 1.1$ & \textbf{88.44} \tiny$\pm 1.27$ & \textbf{89.83} \tiny$\pm 0.8$ \\
        \textbf{MixText} & 32.73 \tiny$\pm 6.02$ & 54.75 \tiny$\pm 3.15$ & 76.59 \tiny$\pm 1.05$ & 82.34 \tiny$\pm 0.94$ \\
        %\textbf{\method(Ours Old)} & 45.47 \tiny$\pm 4.13$ & 58.61 \tiny$\pm 5.06$ & 74.44 \tiny$\pm 2.28$ & 82.29 \tiny$\pm 0.6$  \\
        \textbf{\method} & \textbf{54.16} \tiny$\pm 4.56$ & \textbf{62.71} \tiny$\pm 2.30$ & \textbf{76.73} \tiny$\pm 01.46$ & \textbf{84.45} \tiny$\pm 0.52$ \\

        \midrule
        \multirow{2}{*} & \multicolumn{4}{c}{\underline{HWU64}} \\
        \textbf{Supervised} & 48.87 \tiny$\pm 1.55$ & 63.88 \tiny$\pm 1.6$ & 74.67 \tiny$\pm 1.91$ & 82.21 \tiny$\pm 1.72$ \\
        \textbf{PL} & 48.46 \tiny$\pm 1.86$ & 64.39 \tiny$\pm 1.66$ & 75.76 \tiny$\pm 1.69$ & 82.49 \tiny$\pm 0.94$  \\
        \textbf{PL-T} & 56.9 \tiny$\pm 1.64$ & 68.29 \tiny$\pm 1.79$ & 76.9 \tiny$\pm 1.1$ & 82.96 \tiny$\pm 1.69$  \\
        \textbf{PL-Flex} &  60.15 \tiny$\pm 3.27$ & 69.87 \tiny$\pm 0.93$ & 77.99 \tiny$\pm 1.4$ & 83.83 \tiny$\pm 1.2$  \\
        \textbf{GAN-BERT} & 33.36 \tiny$\pm 16.55$ & 32.9 \tiny$\pm 29.07$ & 72.32 \tiny$\pm 1.41$ & 81.78 \tiny$\pm 1.64$  \\
        \textbf{MixText} & 33.3 \tiny$\pm 8.98$ & 56.46 \tiny$\pm 11.08$ & 66.65 \tiny$\pm 7.28$ & 79.72 \tiny$\pm 1.27$ \\
        %\textbf{\method (Ours Old)} &  62.17 \tiny$\pm 2.67$ & 70.63 \tiny$\pm 1.43$ & 76.99 \tiny$\pm 1.37$ & 82.84 \tiny$\pm 1.04$ \\
        \textbf{\method } &  \textbf{65.33} \tiny$\pm 2.29$ & \textbf{73.03} \tiny$\pm 1.31$ & \textbf{79.63} \tiny$\pm 0.56$ & \textbf{84.59} \tiny$\pm 0.58$ \\
        
        \bottomrule
    \end{tabular}
    }
   %}
    \caption{Mean test accuracy results and their 95\% confidence intervals across 5 repetitions with different different random seeds. All experiments used $k=6$ and $\beta=0.75$. TK-KNN outperformed existing state of the art models, especially when the label set is small. The confidence interval also shows that the TK-KNN results were stable across repetitions.}
    \label{tab:results}
\end{table}
\setlength{\tabcolsep}{6pt}

Results from these experiments are shown in Table~\ref{tab:results}. 
These quantitative results demonstrate that TK-KNN yielded the best performance on the benchmark datasets. We observed the most significant performance gains for CLINC150 and BANKING77, where these datasets have more classes. For instance, on the CLINC150 dataset with 1\% labeled data, our method performs \textbf{10.92\%} better than the second best strategy, FlexMatch. As the portion of labeled data used increases, we notice that the effectiveness of TK-KNN diminishes. 

Another observation from these results is that the GAN-BERT model tends to be unstable when the labeled data is limited. However, GAN-BERT does improve as the proportion of labeled data increases. This is consistent with previous findings on training GANs from computer vision \citep{salimans2016improved, arjovsky2017towards}. We also find that while the MixText method shows improvements the benefits of consistency regularization are not as strong compared to works from the computer vision domain.

These results demonstrate the benefits of TK-KNN's balanced sampling strategy and its use of the distances in the latent space.

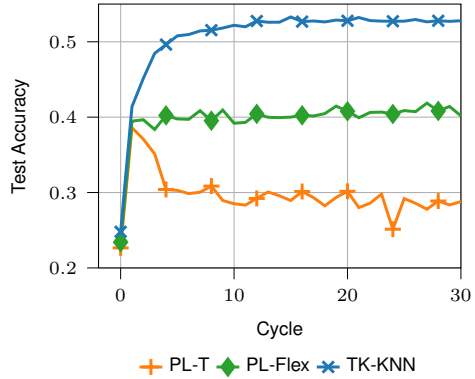
\begin{figure}
    \centering
    % This file was created with tikzplotlib v0.10.1.
\begin{tikzpicture}

\begin{axis}[
height=2.0in,
width=2.5in,
legend style={at={(0.5,-0.3)},anchor=north, font=\sffamily\scriptsize, draw=none},
legend cell align=left,
legend columns=3,
tick align=outside,
tick pos=left,
x grid style={darkgray176},
xlabel={Cycle},
xmajorgrids,
xmin=-1.95, xmax=30,
xtick style={color=black},
y grid style={darkgray176},
ylabel={Test Accuracy},
ymajorgrids,
ymin=0.2, ymax=0.55,
ytick style={color=black}
]

\addplot [line width=1.12pt, darkorange25512714, mark=+, mark size=3, mark repeat=4, mark options={solid}]
table {%
0 0.22658181818181822
1 0.3860363636363636
2 0.37079999999999996
3 0.3513090909090909
4 0.30407272727272733
5 0.3028727272727273
6 0.29858181818181817
7 0.3004
8 0.30854545454545457
9 0.28945454545454546
10 0.28512727272727273
11 0.28334545454545457
12 0.2921818181818182
13 0.3008
14 0.29567272727272725
15 0.28941818181818185
16 0.30138181818181814
17 0.2933090909090909
18 0.28236363636363637
19 0.29356363636363636
20 0.30163636363636365
21 0.28
22 0.2859272727272727
23 0.2979636363636363
24 0.25134545454545454
25 0.29203636363636365
26 0.28578181818181814
27 0.27785454545454547
28 0.28876363636363633
29 0.28356363636363635
30 0.288
};
\addlegendentry{PL-T}
\addplot [line width=1.12pt, forestgreen4416044, mark=diamond*, mark size=3, mark repeat=4, mark options={solid}]
table {%
0 0.23421818181818183
1 0.3946545454545455
2 0.3966181818181818
3 0.38360000000000005
4 0.4023636363636364
5 0.3976363636363636
6 0.3971272727272727
7 0.40858181818181816
8 0.3952727272727273
9 0.4094181818181818
10 0.39192727272727274
11 0.3930909090909091
12 0.40432727272727276
13 0.3998181818181818
14 0.39938181818181817
15 0.40003636363636363
16 0.4024727272727272
17 0.4015272727272727
18 0.4046909090909091
19 0.4146545454545455
20 0.4078181818181818
21 0.3992727272727273
22 0.4062545454545455
23 0.4067272727272727
24 0.40414545454545453
25 0.40869090909090905
26 0.40727272727272723
27 0.4185818181818182
28 0.40825454545454554
29 0.41418181818181815
30 0.4017090909090909
};
\addlegendentry{PL-Flex}

\addplot [line width=1.12pt, steelblue, mark=x, mark size=3, mark repeat=4, mark options={solid}]
table {%
0 0.24789090909090908
1 0.4139636363636364
2 0.45141818181818183
3 0.48465454545454545
4 0.4962545454545454
5 0.5078545454545453
6 0.5096
7 0.5143636363636365
8 0.5153818181818182
9 0.5182181818181818
10 0.5218909090909091
11 0.5198545454545455
12 0.5275272727272726
13 0.5258909090909091
14 0.5259636363636364
15 0.5329090909090909
16 0.526690909090909
17 0.5276
18 0.5263272727272728
19 0.5287636363636364
20 0.5278545454545456
21 0.5321818181818182
22 0.5280727272727272
23 0.5273818181818182
24 0.5272363636363637
25 0.527309090909091
26 0.5295272727272728
27 0.5264
28 0.5278909090909091
29 0.5270545454545454
30 0.5278909090909091
%0 0.22658181818181822
%1 0.37058181818181823
% 2 0.4188727272727273
% 3 0.4449454545454546
% 4 0.45999999999999996
% 5 0.4664727272727272
% 6 0.47000000000000003
% 7 0.4841454545454546
% 8 0.48319999999999996
% 9 0.4825454545454545
% 10 0.4905090909090909
% 11 0.4907636363636364
% 12 0.4870181818181819
% 13 0.4893090909090909
% 14 0.4915272727272727
% 15 0.4917454545454545
% 16 0.48840000000000006
% 17 0.4856727272727273
% 18 0.4813090909090909
% 19 0.48440000000000005
% 20 0.48814545454545455
% 21 0.48563636363636364
% 22 0.4883272727272727
% 23 0.48501818181818174
% 24 0.4899272727272728
% 25 0.48200000000000004
% 26 0.47432727272727265
% 27 0.4808363636363636
% 28 0.4854181818181818
% 29 0.4854181818181818
% 30 0.48509090909090913
};
\addlegendentry{TK-KNN}
\end{axis}

\end{tikzpicture}
    \caption{Convergence analysis of pseudo-labelling strategies on CLINC150 at 1\% labeled data. TK-KNN clearly outperforms the other pseudo-labelling strategies by balancing class pseudo labels after each training cycle. }
    \label{fig:convergence}
\end{figure}

\subsection{Overconfidence in Pseudo-Labelling Regimes}
A key observation we found throughout self-training was that the performance of existing pseudo-labelling methods tended to degrade as the number of cycles increased. An example of this is illustrated in Figure~\ref{fig:convergence}. Here we see that when a pre-defined threshold is used, the model tends to improve performance for the first few training cycles. After that point, the pseudo-labeling becomes heavily biased towards the easier classes. This causes the model to become overconfident in predictions for those classes and neglect more difficult classes. PL-Flex corrects this issue but converges much earlier in the learning process. TK-KNN achieves the best performance thanks to the slower balanced pseudo-labeling approach. This process helps the model learn clearer decision boundaries for all classes simultaneously and prevent overconfidence in the model in some classes.

\subsection{Ablation Study}

Because TK-KNN is different from existing methods in two distinct ways: (1) top-k balanced sampling and (2) KNN ranking, we perform a set of ablation experiments to better understand how each of these affects performance. Specifically, we test TK-KNN under three scenarios, top-k sampling without balancing the classes, top-k sampling with balanced classes, and top-k KNN without balancing for classes. When we perform top-k sampling in an unbalanced manner, we ensure that the total data sampled is still equal to $k * C$, where $C$ is the number of classes. We report the results for these experiments in Table~\ref{tab:ablation}. We also conduct experiments on the addition of the different loss objectives detailed in Appendix~\ref{app:ablation}

%Because TK-KNN is different from existing methods in two distinct ways: (1) top-k balanced sampling and (2) KNN ranking, we perform a set of ablation experiments to better understand how each of these affects performance. Specifically, we test TK-KNN for five different parts top-k sampling, KNN-ranking, contrastive loss, and differential entropy regularization. We show results when our method has each of these parts added and missing. When we peform top-k sampling in an unbalanced manner, we ensure that the total data sampled is still equal to $k * C$, where $C$ is the number of classes. We report the mean test accuracy results and their 95\% confidence intervals for these experiments in Table~\ref{tab:ablation}.

The results from the ablation study demonstrate both the effectiveness of top-k sampling and KNN ranking. A comparison between our unbalanced sampling top-k sampling and balanced versions show a drastic difference in performance across all datasets. We highlight again that the performance difference is greatest in the lowest resource setting, with a \textbf{12.47\%} increase in accuracy for CLINC150 in the 1\% setting.

Results from the TK-KNN method with unbalanced sampling also show an improvement over unbalanced sampling alone. This increase in performance is smaller than the difference between unbalanced and balanced sampling but still highlights the benefits of leveraging the geometry for selective pseudo-labeling. We also present an ablation of the Top-k sampling methodology when passing correctly labeled examples in the Appendix~\ref{app:upper-bound}.

\begin{table}[]
    \centering
    %\small
    \scriptsize{
    %\resizebox{\linewidth}{!}{
    \begin{tabular}{l| ll}
        \toprule
        \multirow{2}{*} &  \multicolumn{2}{c}{\textbf{Percent Labeled}} \\
        \textbf{Method} & \multicolumn{1}{c}{1\%} & \multicolumn{1}{c}{2\%}  \\ 
        \midrule
        \multirow{2}{*} & \multicolumn{2}{c}{\underline{CLINC150}} \\
        \textbf{Top-k U} & 38.37 \tiny$\pm 1.08$ & 55.0 \tiny$\pm 1.44$ \\
        \textbf{Top-k B} & 51.36 \tiny$\pm 2.1$ & 64.99 \tiny$\pm 0.64$ \\
        \textbf{Top-k KNN U} & 41.24 \tiny$\pm 0.97$ & 55.01 \tiny$\pm 1.49$ \\
        \textbf{Top-k KNN B} & \textbf{53.73} \tiny$\pm 1.72$ & \textbf{65.87} \tiny$\pm 1.18$ \\
        
        \midrule
        \multirow{2}{*} & \multicolumn{2}{c}{\underline{BANKING77}} \\
        \textbf{Top-k U} & 41.56 \tiny$\pm 4.73$ & 54.78 \tiny$\pm 3.51$ \\
        \textbf{Top-k B} & 50.45 \tiny$\pm 4.53$ & 63.19 \tiny$\pm 1.78$ \\
        \textbf{Top-k KNN U} & 44.12 \tiny$\pm 3.14$ & 55.9 \tiny$\pm 2.65$ \\
        \textbf{Top-k KNN B} & \textbf{54.16} \tiny$\pm 4.56$ & \textbf{62.71} \tiny$\pm 2.30$ \\
        \midrule
        \multirow{2}{*} & \multicolumn{2}{c}{\underline{HWU64}} \\
        \textbf{Top-k U} & 54.87 \tiny$\pm 1.64$ & 64.85 \tiny$\pm 1.54$ \\
        \textbf{Top-k B} & 54.13 \tiny$\pm 6.0$ & 65.12 \tiny$\pm 0.35$ \\
        \textbf{Top-k KNN U} & 57.86 \tiny$\pm 2.25$ & 69.33 \tiny$\pm 0.96$ \\
        \textbf{Top-k KNN B} & \textbf{65.33} \tiny$\pm 2.29$ & \textbf{73.03} \tiny$\pm 1.31$ \\

        \bottomrule
    \end{tabular}
    }
    \caption{Ablation study of top-k sampling. U stands for unbalanced sampling, where classes are not balanced. B is for balanced sampling, and classes are balanced with the top-k per class.}
    \label{tab:ablation}
\end{table}

\begin{figure}[t]
%\begin{subfigure}{.45\textwidth}
    \centering
    % This file was created with tikzplotlib v0.10.1.
\begin{tikzpicture}

\definecolor{darkgray176}{RGB}{176,176,176}
\definecolor{darkorange25512714}{RGB}{255,127,14}
\definecolor{lightgray204}{RGB}{204,204,204}
\definecolor{steelblue31119180}{RGB}{31,119,180}

\begin{axis}[
height=2.0in,
width=2.75in,
legend style={at={(0.5,-0.25)},anchor=north, font=\sffamily\scriptsize, draw=none},
legend cell align=left,
legend columns=5,
tick align=outside,
tick pos=left,
x grid style={darkgray176},
xlabel={Cycle},
xmajorgrids,
xmin=-1.95, xmax=30,
xtick style={color=black},
y grid style={darkgray176},
ylabel={Test Accuracy},
ymajorgrids,
ymin=0.485, ymax=0.655,
ytick style={color=black}
]

\addplot[line width=1.12pt, steelblue31119180, mark=+, mark size=3, mark repeat=4, mark options={solid}]
table {%
0 0.49535315985130113
1 0.583457249070632
2 0.6037174721189591
3 0.612639405204461
4 0.6278810408921933
5 0.6317843866171005
6 0.6239776951672862
7 0.6278810408921933
8 0.6302973977695168
9 0.6306691449814127
10 0.6258364312267658
11 0.6256505576208178
12 0.6278810408921933
13 0.6221189591078067
14 0.6228624535315985
15 0.616542750929368
16 0.6223048327137546
17 0.6263940520446096
18 0.6256505576208178
19 0.6241635687732341
20 0.6269516728624536
21 0.6232342007434944
22 0.6226765799256505
23 0.6234200743494424
24 0.6263940520446096
25 0.6243494423791822
26 0.6230483271375464
27 0.6254646840148699
28 0.6245353159851301
29 0.6250929368029741
30 0.6243494423791822
};\addlegendentry{0.0}

\addplot[line width=1.12pt, magenta, mark=square, mark size=3, mark repeat=4, mark options={solid}]
table{%
0 0.49535315985130113
1 0.587918215613383
2 0.6033457249070633
3 0.6213754646840148
4 0.6286245353159852
5 0.6325278810408921
6 0.6425650557620818
7 0.6390334572490706
8 0.6394052044609666
9 0.6355018587360595
10 0.6388475836431227
11 0.637546468401487
12 0.6356877323420074
13 0.629368029739777
14 0.63364312267658
15 0.6369888475836432
16 0.6301115241635687
17 0.6332713754646841
18 0.6362453531598513
19 0.6347583643122676
20 0.641449814126394
21 0.6345724907063197
22 0.6304832713754647
23 0.6321561338289963
24 0.6327137546468401
25 0.6319702602230484
26 0.6349442379182155
27 0.6310408921933085
28 0.6336431226765799
29 0.6327137546468402
30 0.6310408921933085
};\addlegendentry{0.25}

\addplot[line width=1.12pt, black, mark=pentagon, mark size=3, mark repeat=4, mark options={solid}]
table{%
0 0.49535315985130113
1 0.5910780669144982
2 0.603903345724907
3 0.6232342007434944
4 0.6276951672862453
5 0.625278810408922
6 0.6254646840148699
7 0.633085501858736
8 0.628996282527881
9 0.6249070631970259
10 0.6271375464684015
11 0.6273234200743494
12 0.6262081784386618
13 0.6319702602230484
14 0.6301115241635687
15 0.6302973977695168
16 0.6327137546468402
17 0.6304832713754647
18 0.6282527881040892
19 0.6278810408921933
20 0.6304832713754648
21 0.6288104089219331
22 0.6314126394052045
23 0.6308550185873606
24 0.628438661710037
25 0.6234200743494424
26 0.6312267657992565
27 0.6278810408921933
28 0.6304832713754647
29 0.6314126394052044
30 0.6301115241635687
};\addlegendentry{0.5}

\addplot [line width=1.12pt, darkorange25512714, mark=diamond*, mark size=3, mark repeat=4, mark options={solid}]
table {%
0 0.49535315985130113
1 0.5955390334572491
2 0.6074349442379182
3 0.6208178438661711
4 0.633457249070632
5 0.642007434944238
6 0.6340148698884759
7 0.6514869888475836
8 0.6397769516728624
9 0.6421933085501859
10 0.6327137546468402
11 0.6362453531598513
12 0.636802973977695
13 0.6364312267657992
14 0.6358736059479554
15 0.6314126394052045
16 0.6325278810408921
17 0.633271375464684
18 0.63364312267658
19 0.6334572490706318
20 0.6317843866171003
21 0.6325278810408922
22 0.6325278810408923
23 0.6343866171003716
24 0.6301115241635687
25 0.625650557620818
26 0.6325278810408921
27 0.6286245353159851
28 0.6323420074349442
29 0.6312267657992565
30 0.6326420074349442
};
\addlegendentry{0.75}
\addplot [line width=1.12pt, forestgreen4416044,  mark=x, mark size=3, mark repeat=4, mark options={solid,rotate=180}]
table {%
0 0.49535315985130113
1 0.5973977695167287
2 0.6079925650557622
3 0.6176579925650557
4 0.629368029739777
5 0.637360594795539
6 0.6256505576208179
7 0.6343866171003717
8 0.6267657992565056
9 0.63364312267658
10 0.625278810408922
11 0.6275092936802974
12 0.6328996282527881
13 0.6310408921933085
14 0.6284386617100373
15 0.6314126394052044
16 0.6275092936802974
17 0.6345724907063197
18 0.6325278810408921
19 0.6342007434944238
20 0.6351301115241635
21 0.6368029739776951
22 0.6386617100371746
23 0.6388475836431227
24 0.6371747211895911
25 0.6390334572490707
26 0.6355018587360595
27 0.637360594795539
28 0.6397769516728624
29 0.64182156133829
30 0.64082156133829
};
\addlegendentry{1.0}
\end{axis}

\end{tikzpicture}  
    % \caption{}
    \label{fig:hwu_beta_ablation}
    \caption{A comparison of TK-KNN on HWU64 with 1\% labeled data as $\beta$ varies.}
    \label{fig:parameter_ablation}
\end{figure}
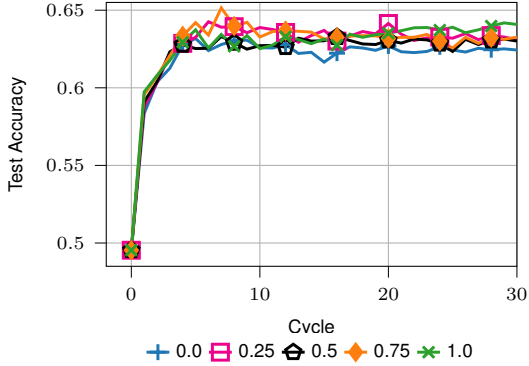

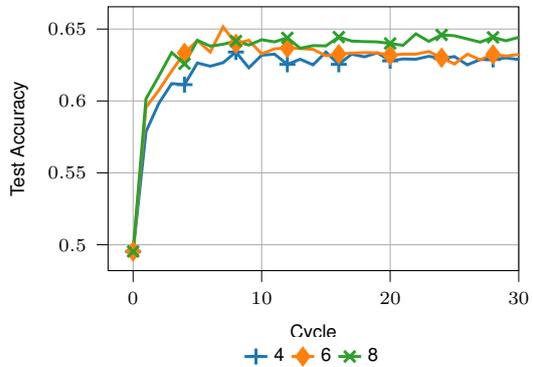
\begin{figure}[t]
%\begin{subfigure}{.45\textwidth}
    \centering
    % This file was created with tikzplotlib v0.10.1.
\begin{tikzpicture}

\definecolor{darkgray176}{RGB}{176,176,176}
\definecolor{darkorange25512714}{RGB}{255,127,14}
\definecolor{forestgreen4416044}{RGB}{44,160,44}
\definecolor{lightgray204}{RGB}{204,204,204}
\definecolor{steelblue31119180}{RGB}{31,119,180}

\begin{axis}[
height=2.0in,
width=2.75in,
legend style={at={(0.5,-0.25)},anchor=north, font=\sffamily\scriptsize, draw=none},
legend cell align=left,
legend columns=3,
tick align=outside,
tick pos=left,
x grid style={darkgray176},
xlabel={Cycle},
ylabel={Test Accuracy},
xmajorgrids,
xmin=-1.95, xmax=30,
xtick style={color=black},
y grid style={darkgray176},
ymajorgrids,
ymin=0.482031191150126, ymax=0.665482918056482,
ytick style={color=black},
%ymajorticks=false,
]

\addplot [line width=1.12pt, steelblue31119180, mark=+, mark size=3, mark repeat=4, mark options={solid,rotate=270}]
table {%
0 0.49535315985130113
1 0.5786245353159851
2 0.5983271375464685
3 0.6120817843866171
4 0.6113382899628252
5 0.6263940520446096
6 0.6241635687732342
7 0.6265799256505575
8 0.6340148698884759
9 0.6230483271375464
10 0.6317843866171003
11 0.6325278810408922
12 0.6254646840148699
13 0.6289962825278811
14 0.625092936802974
15 0.6340148698884759
16 0.6254646840148699
17 0.6325278810408922
18 0.6306691449814127
19 0.633457249070632
20 0.6278810408921933
21 0.629182156133829
22 0.628996282527881
23 0.6310408921933085
24 0.6295539033457249
25 0.6308550185873607
26 0.6250929368029741
27 0.6288104089219331
28 0.6288104089219331
29 0.6299256505576208
30 0.6288104089219331
};
\addlegendentry{4}
\addplot [line width=1.12pt, darkorange25512714, mark=diamond*, mark size=3, mark repeat=4, mark options={solid}]
table {%
0 0.49535315985130113
1 0.5955390334572491
2 0.6074349442379182
3 0.6208178438661711
4 0.633457249070632
5 0.642007434944238
6 0.6340148698884759
7 0.6514869888475836
8 0.6397769516728624
9 0.6421933085501859
10 0.6327137546468402
11 0.6362453531598513
12 0.636802973977695
13 0.6364312267657992
14 0.6358736059479554
15 0.6314126394052045
16 0.6325278810408921
17 0.633271375464684
18 0.63364312267658
19 0.6334572490706318
20 0.6317843866171003
21 0.6325278810408922
22 0.6325278810408923
23 0.6343866171003716
24 0.6301115241635687
25 0.625650557620818
26 0.6325278810408921
27 0.6286245353159851
28 0.6323420074349442
29 0.6312267657992565
30 0.6323420074349442
};
\addlegendentry{6}
\addplot [line width=1.12pt, forestgreen4416044, mark=x, mark size=3, mark repeat=4, mark options={solid}]
table {%
0 0.49535315985130113
1 0.6014869888475837
2 0.6172862453531598
3 0.63364312267658
4 0.6258364312267657
5 0.6421933085501859
6 0.6381040892193309
7 0.6394052044609665
8 0.641635687732342
9 0.6388475836431227
10 0.6425650557620818
11 0.6410780669144981
12 0.6438661710037175
13 0.6366171003717472
14 0.6384758364312267
15 0.6382899628252788
16 0.6444237918215613
17 0.641635687732342
18 0.6412639405204461
19 0.6410780669144981
20 0.6399628252788104
21 0.6386617100371746
22 0.6466542750929367
23 0.641449814126394
24 0.645910780669145
25 0.6453531598513012
26 0.6431226765799256
27 0.6408921933085502
28 0.6442379182156134
29 0.6418215613382899
30 0.6442379182156134
};
\addlegendentry{8}
\end{axis}
\end{tikzpicture}  
    % \caption{}
    \label{fig:hwu_top_k_ablation}
    \caption{A comparison of TK-KNN on HWU64 with 1\% labeled data as k varies.}
    \label{fig:top_k_ablation}
\end{figure}

%The results of our top-k sampling and it's variant demonstrate that performance improves when we utilize K-nearest neighbors search during pseudo labeling. The KNN search helps our method by utilizing the smoothness assumption \cite{ouali2020overview}, that states if points lie in a high-density region are similar, then their outputs should be similar.

%We perform an ablation study to determine the variations in performance that result from using various values for $k$ and $\beta$.

%\subsection{Upper Bound Analysis}
%We further ran experiments to gauge the performance of top-k sampling when ground truth labels are fed to the model instead of predicted pseudo labels. This experiment gives us an indicator as to how performance should increase throughout the self-training process in an ideal pseudo-labeling scenario.
%We present the results of this in Figure.~\ref{fig:upper_bound}. As expected, the model tends to converge towards a fully supervised performance as the cycle increases and more data is (pseudo-)labeled. Another point of interest is that the method's upper bound can continue learning with proper labels, while TK-KNN method tends to converge earlier. The upper bound method also takes a significant increase in the first few cycles as well. This highlights a need to investigate methods for accurate pseudo label selection further, so that the model can continue to improve.

\subsection{Parameter Search}
TK-KNN relies on two hyperparameters $k$ and $\beta$ that can affect performance based on how they are configured. We explore experiments to gauge their effect on learning by testing $k \in (4,6,8)$ and $\beta \in (0.0, 0.25, 0.50, 0.75, 1.00)$. When varying $k$ we hold $\beta$ at 0.75. For $\beta$ experiments we keep $k=6$. When $\beta = 0.0$, this is equivalent to just top-k sampling based on Equation \ref{eq:2}. Alternatively, when $\beta = 1.0$, this is equivalent to only using the KNN similarity for ranking.
Results from our experiments are shown in Figures~\ref{fig:parameter_ablation} for $\beta$ and~\ref{fig:top_k_ablation} for $k$.

As we varied the $\beta$ parameter, we noticed that all configurations tended to have similar training patterns. After we trained the model for the first five cycles, the model tended to move in small jumps between subsequent cycles. From the illustration, we can see that no single method was always the best, but the model tended to perform worse when $\beta = 0.0$, highlighting the benefits of including our KNN similarity for ranking. The model reached the best performance when $\beta = 0.75$, which occurs about a third of the way through the training process.

Comparison of values for $k$ show that \method is robust to adjustments in this hyperparameter. We notice slight performance benefits from selecting a higher $k$ of 6 and 8 in comparison to 4. When a higher value of $k$ is used the model will see an increase in performance earlier in the self-training process, as it has more examples to train from. This is only acheivable though when high quality correct samples are selected across the entire class distribution. If a $k$ value was selected that is too large, more bad examples will be included early in the training process and may result in poor model performance.

%Comparison of values for $k$ show more consistent performance for $k=6$ than our other two settings.
%Throughout almost the entire training process, $k=6$ results in better performance.
%We also see that when a larger $k$ is used, this can lead to too many bad examples being included in the training process early on, which results in worse performance.

\begin{comment}
\paragraph{Choosing $k$}
If one were to chose $k=1$ then more cycles of self-training may be necessary to allow size of sampled pseudo labels to increase in accordance with the dataset. As a lower bound, one can chose to train the model with the smallest amount of data per cycle to ensure the entire dataset gets used for self-training via the following equation.
\begin{equation}
    k = 
\end{equation}
\end{comment}

\section{Conclusions}
This paper introduces TK-KNN, a balanced distance-based pseudo-labeling approach for semi-supervised intent classification. TK-KNN deviates from previous pseudo-labeling methods as it does not rely on a threshold to select the samples. Instead, we show that a balanced approach that takes the model prediction and K-Nearest Neighbor similarity measure allows for more robust decision boundaries to be learned. Experiments on three popular intent classification datasets, CLINC150, Banking77, and Hwu64, demonstrate that our method improved performance in all scenarios.

%DO NOT DELETE THIS SECTION - MANDATORY REQUIREMENT FOR ACL 2023.
%This section does not count towards the page limit.
\section{Limitations}
While our method shows noticeable improvements, it is not without limitations. Our method does not require searching for a good threshold but instead requires two different hyperparameters, $k$ and $\beta$, that must be found. We offer a reasonable method and findings for selecting both of these but others may want to search for other combinations depending on the dataset. A noticeable drawback from our self-training method is that more cycles of training will need to be done, especially if the value of $k$ is small. This requires much more GPU usage to converge to a good point. Further, we did not explore any heavily imbalanced datasets, so we are unaware of how \method would perform under those scenarios.

%Does not count towards page limit
\section{Ethics Statement}
This work can be used to help improve current virtual assistant systems for many of the businesses that may be seeking to use one. As the goal of these systems is to understand a persons request, failures can lead to wrong actions being taken that potentially impact an individual.

{\small
\bibliographystyle{abbrvnat}
\bibliography{refs}
}

\appendix

%\section{Imbalanced Dataset Experiments}
%\label{app:imbalanced}
%As our method relies on a balanced sampling method a question naturally arises as to how it will perform on an unbalanced dataset. Here we present results on the TREC dataset \citep{hovy2001trec}. TREC is a question dataset containing labeled questions. The dataset contains coarse-grained labels and fine-grained labels. The dataset is imbalanced for both of these label classes but we evaluate on the fine-grained labels as this further restricts the imbalanced class size for the model. For our results in this experiment we report accuracy, macro-recall, marco-precision, and macro-f1 score.

\begin{algorithm}
\caption{\method Sampling For a Cycle}\label{alg:cap}
\begin{algorithmic}[1]

\Require Data of $X = {x_n, y_n}: n \in (1,...,N)$, $U = {u_m}: m \in (1,...,M)$, $\beta$
%\Ensure Learned model parameters $\theta$

\State Predict pseudo-labels for all $U$
\State Calculate cosine similarity via. Eq. (2) for all $U$ per class
\State Calculate score via Eq.(3)
\State Combine $X$ and top-k per class from $U$
\end{algorithmic}
\end{algorithm}

\section{Upper Bound Analysis}
\label{app:upper-bound}
\begin{figure*}[t]
\begin{subfigure}{.33\textwidth}
    \centering
    % This file was created with tikzplotlib v0.10.1.
\begin{tikzpicture}

\begin{axis}[
height=1.75in,
width=2.25in,
legend cell align={left},
legend style={
  fill opacity=0.8,
  draw opacity=1,
  text opacity=1,
  at={(5.05,1)},
  anchor=north,
  legend columns=3
},
legend to name={mylegend},
tick align=outside,
tick pos=left,
x grid style={darkgray176},
xlabel={Cycle},
title={CLINC150},
xmajorgrids,
xmin=-1.95, xmax=30,
xtick style={color=black},
y grid style={darkgray176},
ylabel={Test Accuracy},
ymajorgrids,
ymin=0.206704116330275, ymax=0.925443174164871,
ytick style={color=black}
]

\addplot [line width=1.12pt, darkorange25512714, mark=+, mark size=3, mark repeat=4, mark options={solid}]
table {%
0 0.24789090909090908
1 0.4139636363636364
2 0.45141818181818183
3 0.48465454545454545
4 0.4962545454545454
5 0.5078545454545453
6 0.5096
7 0.5143636363636365
8 0.5153818181818182
9 0.5182181818181818
10 0.5218909090909091
11 0.5198545454545455
12 0.5275272727272726
13 0.5258909090909091
14 0.5259636363636364
15 0.5329090909090909
16 0.526690909090909
17 0.5276
18 0.5263272727272728
19 0.5287636363636364
20 0.5278545454545456
21 0.5321818181818182
22 0.5280727272727272
23 0.5273818181818182
24 0.5272363636363637
25 0.527309090909091
26 0.5295272727272728
27 0.5264
28 0.5278909090909091
29 0.5270545454545454
30 0.5278909090909091
};
\addlegendentry{\method}
\addplot [line width=1.12pt, forestgreen4416044, mark=diamond*, mark size=3, mark repeat=4, mark options={solid}]
table {%
0 0.21680000000000002
1 0.5786545454545454
2 0.6276363636363637
3 0.6544363636363636
4 0.6730909090909091
5 0.6915272727272727
6 0.71
7 0.7174545454545453
8 0.7322545454545455
9 0.7540727272727274
10 0.7633818181818182
11 0.7781818181818181
12 0.7963272727272728
13 0.8041818181818183
14 0.8065818181818182
15 0.8138181818181817
16 0.8213818181818182
17 0.8330909090909092
18 0.8314909090909091
19 0.8370181818181818
20 0.8296727272727272
21 0.8312727272727273
22 0.839709090909091
23 0.8369818181818183
24 0.8422545454545454
25 0.8355272727272727
26 0.841090909090909
27 0.8484
28 0.8335272727272727
29 0.8483636363636362
30 0.852290909090909
};
\addlegendentry{Top-K Upper Bound}
\end{axis}

\end{tikzpicture}  
%    \caption{CLINC}
    \label{fig:clinc_ablation}
\end{subfigure}
\hspace{0.5cm}
\begin{subfigure}{.25\textwidth}
   \centering
    % This file was created with tikzplotlib v0.10.1.
\begin{tikzpicture}

\definecolor{darkgray176}{RGB}{176,176,176}
\definecolor{darkorange25512714}{RGB}{255,127,14}
\definecolor{forestgreen4416044}{RGB}{44,160,44}
\definecolor{lightgray204}{RGB}{204,204,204}
\definecolor{steelblue31119180}{RGB}{31,119,180}

\begin{axis}[
height=1.75in,
width=2.25in,
legend cell align={left},
legend style={
  fill opacity=0.8,
  draw opacity=1,
  text opacity=1,
  at={(1.05,1)},
  anchor=north west,
  draw=lightgray204
},
tick align=outside,
tick pos=left,
x grid style={darkgray176},
xlabel={Cycle},
title={BANKING77},
xmajorgrids,
xmin=-1.95, xmax=30,
xtick style={color=black},
y grid style={darkgray176},
ylabel={},
ymajorgrids,
ymin=0.206704116330275, ymax=0.975443174164871,
ymajorticks=false,
%ytick style={color=black}
]

\addplot [line width=1.12pt, darkorange25512714, mark=+, mark size=3, mark repeat=4, mark options={solid}]
table {%
0 0.340064935064935
1 0.47740259740259744
2 0.5105844155844156
3 0.5175324675324675
4 0.5178571428571429
5 0.5184415584415584
6 0.5305194805194805
7 0.5279220779220779
8 0.5303246753246753
9 0.5301948051948052
10 0.5272077922077922
11 0.5260389610389611
12 0.5269480519480519
13 0.5249350649350649
14 0.5259090909090909
15 0.5242857142857142
16 0.5252597402597403
17 0.518896103896104
18 0.5168181818181818
19 0.5112987012987013
20 0.49961038961038956
21 0.49467532467532466
22 0.49512987012987014
23 0.49577922077922076
24 0.4974025974025974
25 0.49331168831168837
26 0.487987012987013
27 0.4993506493506493
28 0.4948701298701299
29 0.4928571428571429
30 0.4948701298701299
};

\addplot [line width=1.12pt, forestgreen4416044, mark=diamond*, mark size=3, mark repeat=4, mark options={solid}]
table {%
0 0.2733766233766234
1 0.6024675324675324
2 0.6546753246753246
3 0.6721428571428572
4 0.6895454545454546
5 0.7033766233766234
6 0.7232467532467532
7 0.7500649350649351
8 0.7495454545454546
9 0.7775974025974026
10 0.7966233766233767
11 0.8195454545454546
12 0.8447402597402597
13 0.8505844155844156
14 0.8668181818181818
15 0.8675974025974027
16 0.8812337662337664
17 0.8856493506493507
18 0.8898701298701299
19 0.8922077922077923
20 0.8992857142857142
21 0.9044155844155843
22 0.9012987012987013
23 0.907077922077922
24 0.9118181818181819
25 0.910974025974026
26 0.9110389610389611
27 0.9150649350649351
28 0.9087662337662337
29 0.9110389610389611
30 0.9104545454545455
};

\end{axis}

\end{tikzpicture}  
    %\caption{BANKING77}
    \label{fig:banking_ablation}
\end{subfigure}
\hspace{0.5cm}
\begin{subfigure}{.25\textwidth}
   \centering
    % This file was created with tikzplotlib v0.10.1.
\begin{tikzpicture}

\definecolor{darkgray176}{RGB}{176,176,176}
\definecolor{darkorange25512714}{RGB}{255,127,14}
\definecolor{forestgreen4416044}{RGB}{44,160,44}
\definecolor{lightgray204}{RGB}{204,204,204}
\definecolor{steelblue31119180}{RGB}{31,119,180}

\begin{axis}[
height=1.75in,
width=2.25in,
legend cell align={left},
legend style={
  fill opacity=0.8,
  draw opacity=1,
  text opacity=1,
  at={(1.05,1)},
  anchor=north west,
  draw=lightgray204
},
tick align=outside,
tick pos=left,
x grid style={darkgray176},
xlabel={Cycle},
title={HWU64},
xmajorgrids,
xmin=-1.95, xmax=30,
xtick style={color=black},
y grid style={darkgray176},
ymajorgrids,
ylabel={},
ymin=0.206704116330275, ymax=0.975443174164871,
ytick style={color=black},
ymajorticks=false,
]

\addplot [line width=1.12pt, darkorange25512714, mark=+, mark size=3, mark repeat=4, mark options={solid}]
table {%
0 0.49535315985130113
1 0.5955390334572491
2 0.6074349442379182
3 0.6208178438661711
4 0.633457249070632
5 0.642007434944238
6 0.6340148698884759
7 0.6514869888475836
8 0.6397769516728624
9 0.6421933085501859
10 0.6327137546468402
11 0.6362453531598513
12 0.636802973977695
13 0.6364312267657992
14 0.6358736059479554
15 0.6314126394052045
16 0.6325278810408921
17 0.633271375464684
18 0.63364312267658
19 0.6334572490706318
20 0.6317843866171003
21 0.6325278810408922
22 0.6325278810408923
23 0.6343866171003716
24 0.6301115241635687
25 0.625650557620818
26 0.6325278810408921
27 0.6286245353159851
28 0.6323420074349442
29 0.6312267657992565
30 0.6312267657992565
};
\addplot [line width=1.12pt, forestgreen4416044, mark=diamond*, mark size=3, mark repeat=4, mark options={solid}]
table {%
0 0.41654275092936804
1 0.6120817843866171
2 0.6773234200743494
3 0.6973977695167287
4 0.7195167286245353
5 0.7328996282527881
6 0.7434944237918215
7 0.7529739776951674
8 0.7576208178438661
9 0.7641263940520446
10 0.7860594795539033
11 0.8031598513011152
12 0.8005576208178438
13 0.812639405204461
14 0.8152416356877324
15 0.820817843866171
16 0.8332713754646841
17 0.8433085501858736
18 0.8492565055762082
19 0.8594795539033457
20 0.8689591078066915
21 0.8773234200743495
22 0.8808550185873605
23 0.8843866171003718
24 0.8890334572490707
25 0.8923791821561338
26 0.8983271375464685
27 0.9007434944237918
28 0.8912639405204461
29 0.8934944237918216
30 0.8955390334572492
};

\end{axis}

\end{tikzpicture}  
    %\caption{HWU64}
    \label{fig:hwu_ablation}
\end{subfigure}
\begin{center}
    \ref{mylegend}
\end{center}
\caption{Ablation results for each dataset using 1\% labeled data. }
\label{fig:upper_bound}
\end{figure*}
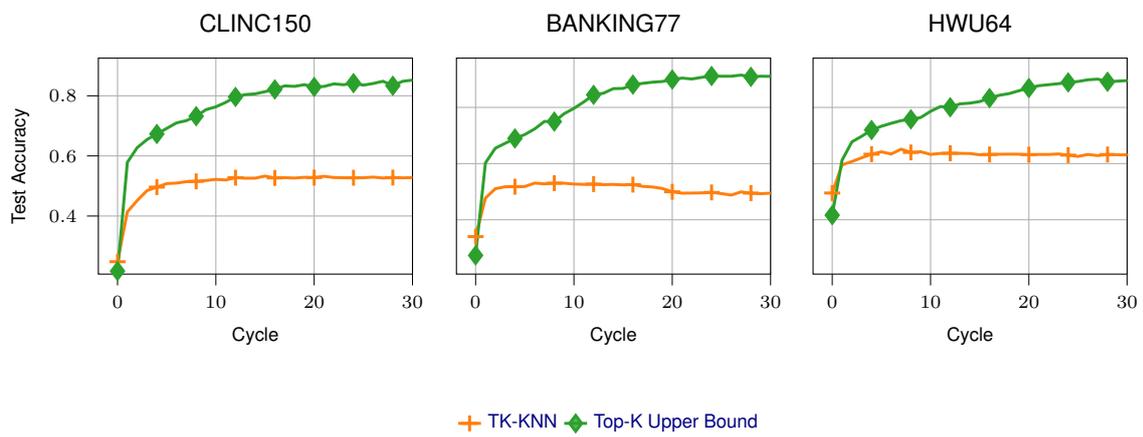

We further ran experiments to gauge the performance of top-k sampling when ground truth labels are fed to the model instead of predicted pseudo labels. This experiment gives us an indicator as to how performance should increase throughout the self-training process in an ideal pseudo-labeling scenario.
We present the results of this in Figure.~\ref{fig:upper_bound}. As expected, the model tends to converge towards a fully supervised performance as the cycle increases and more data is (pseudo-)labeled. Another point of interest is that the method's upper bound can continue learning with proper labels, while TK-KNN method tends to converge earlier. The upper bound method also takes a significant increase in the first few cycles as well. This highlights a need to investigate methods for accurate pseudo label selection further, so that the model can continue to improve.

\section{MixText Experiments}
\label{app:mix-text}
The MixText method requires a number of hyperparameters to be selected in order to acheive good performance. The method also relies on data augmentation of the original text. For our experiments we used the \texttt{Helsinki-NLP} models available on the huggingface repository \footnote{\footnotesize https://huggingface.co/Helsinki-NLP/opus-mt-en-de https://huggingface.co/Helsinki-NLP/opus-mt-en-ru} to perform back translation. Following the original paper we performed back translation into German (de) and Russian (ru). Our experiments used a labeled batch size of 4 and unlabeled batch size of 8, the same as the original paper. The Temperature parameter was set to 0.5.

\section{Ablation Table}
\label{app:ablation}
We present a detailed breakdown of our full ablation results in Table~\ref{tab:full_ablation}. These results include the performance when using only the cross-entropy loss, as well as various combinations of the supervised contrastive loss and differential entropy regularizer. The results in this table demonstrate that the inclusion of additional loss objectives improves performance. This is particularly evident when we add our KNN objective, as we observe an increase of approximately 1-4\%. Although the addition of the differential entropy regularizer yields smaller performance improvements, it remains beneficial to our method overall.

\begin{table}[]
    \centering
    \resizebox{\linewidth}{!}{
    \begin{tabular}{l| cccc}
        \toprule
        \multirow{2}{*} &  \multicolumn{2}{c}{\textbf{Percent Labeled}} \\
        \textbf{Method} & 1\% & 2\%  \\ 
        \midrule
        \multirow{2}{*} & \multicolumn{2}{c}{\underline{CLINC150}} \\
        \textbf{Top-k U CE} & 37.06 \small$\pm 2.09$ & 50.8 \small$\pm 1.79$  \\
        \textbf{Top-k U CE CON} & 38.67 \small$\pm 1.47$ & 55.09 \small$\pm 0.85$ \\
        %\textbf{Top-k U CE DER} & 28.76 \small$\pm 1.95$ & 48.22 \small$\pm 1.49$ \\
        \textbf{Top-k U CE CON DER} & 38.37 \small$\pm 1.08$ & 55.0 \small$\pm 1.44$ \\
        \textbf{Top-k B CE} & 49.53 \small$\pm 3.1$ & 62.45 \small$\pm 1.21$ \\
        \textbf{Top-k B CE CON} & 52.33 \small$\pm 2.53$ & 64.52 \small$\pm 1.63$ \\
        \textbf{Top-k B CE CON DER} & 51.36 \small$\pm 2.1$ & 64.99 \small$\pm 0.64$ \\
        \textbf{Top-k KNN U CE} & 40.58 \small$\pm 2.54$ & 53.45 \small$\pm 1.05$  \\
        \textbf{Top-k KNN U CE CON} & 42.3 \small$\pm 2.86$ & 56.07 \small$\pm 1.26$ \\
        \textbf{Top-k KNN U CE CON DER} & 41.24 \small$\pm 0.97$ & 55.01 \small$\pm 1.49$ \\
        \textbf{Top-k KNN B CE} & 50.28 \small$\pm 3.00$ & 63.99 \small$\pm 0.85$ \\
        \textbf{Top-k KNN B CE CON} & 53.23 \small$\pm 3.17$ & 65.37 \small$\pm 1.17$ \\
        \textbf{Top-k KNN B CE CON DER} & \textbf{53.73} \small$\pm 1.72$ & \textbf{65.87} \small$\pm 1.18$ \\
        
        \midrule
        \multirow{2}{*} & \multicolumn{2}{c}{\underline{BANKING77}} \\
        \textbf{Top-k U CE} & 36.19 \small$\pm 2.93$ & 51.69 \small$\pm 3.54$ \\
        \textbf{Top-k U CE CON} & 42.16 \small$\pm 4.61$ & 54.77 \small$\pm 2.48$ \\
        %\textbf{Top-k U CE DER} & 33.66 \small$\pm 3.71$ & 47.67 \small$\pm 1.81$ \\
        \textbf{Top-k U CE CON DER} & 41.56 \small$\pm 4.73$ & 54.78 \small$\pm 3.51$ \\
        \textbf{Top-k B CE} & 44.35 \small$\pm 3.82$ & 57.19 \small$\pm 3.23$ \\
        \textbf{Top-k B CE CON} & 50.48 \small$\pm 3.75$ & 62.34 \small$\pm 1.57$ \\
        \textbf{Top-k B CE CON DER} & 50.45 \small$\pm 4.53$ & \textbf{63.19} \small$\pm 1.78$ \\
        \textbf{Top-k KNN U CE} & 36.96 \small$\pm 4.78$ & 52.42 \small$\pm 2.75$  \\
        \textbf{Top-k KNN U CE CON} & 44.86 \small$\pm 3.53$ & 56.49 \small$\pm 2.68$ \\
        \textbf{Top-k KNN U CE CON DER} & 44.12 \small$\pm 3.14$ & 55.9 \small$\pm 2.65$ \\
        \textbf{Top-k KNN B CE} & 49.16 \small$\pm 3.02$ & 59.81 \small$\pm 1.69$ \\
        \textbf{Top-k KNN B CE CON} & 52.4 \small$\pm 3.91$ & 62.35 \small$\pm 2.18$ \\
        \textbf{Top-k KNN B CE CON DER} & \textbf{54.16} \small$\pm 4.56$ & 62.71 \small$\pm 2.30$ \\

        \midrule
        \multirow{2}{*} & \multicolumn{2}{c}{\underline{HWU64}} \\
        \textbf{Top-k U CE} &  51.45 \small$\pm 3.8$ & 62.79 \small$\pm 1.99$   \\
        \textbf{Top-k U CE CON} & 54.8 \small$\pm 2.1$ & 66.58 \small$\pm 1.17$ \\
        %\textbf{Top-k U CE DER} & 46.6 \small$\pm 1.43$ & 59.48 \small$\pm 1.7$ \\
        \textbf{Top-k U CE CON DER} & 54.87 \small$\pm 1.64$ & 64.85 \small$\pm 1.54$ \\
        \textbf{Top-k B CE} &  61.69 \small$\pm 3.15$ & 70.41 \small$\pm 0.66$  \\
        \textbf{Top-k B CE CON} & 63.05 \small$\pm 2.6$ & 72.55 \small$\pm 1.41$ \\
        \textbf{Top-k B CE CON DER} & 54.13 \small$\pm 6.0$ & 65.12 \small$\pm 0.35$ \\
        \textbf{Top-k KNN U CE} & 52.97 \small$\pm 1.45$ & 64.41 \small$\pm 1.58$ \\
        \textbf{Top-k KNN U CE CON} & 59.26 \small$\pm 2.99$ & 69.98 \small$\pm 1.44$ \\
        \textbf{Top-k KNN U CE CON DER} & 57.86 \small$\pm 2.25$ & 69.33 \small$\pm 0.96$ \\
        \textbf{Top-k KNN B CE} & 62.43 \small$\pm 2.78$ & 71.04 \small$\pm 0.93$ \\
        \textbf{Top-k KNN B CE CON} &  64.41 \small$\pm 2.26$ & 72.71 \small$\pm 1.07$ \\
        \textbf{Top-k KNN B CE CON DER} & \textbf{65.33} \small$\pm 2.29$ & \textbf{73.03} \small$\pm 1.31$ \\

        \bottomrule
    \end{tabular}
    }
    \caption{Ablation study of top-k sampling. U stands for unbalanced sampling, where classes are not balanced. B is for balanced sampling, and classes are balanced with the top-k per class. CE stands for cross-entropy loss, CON for contrastive loss, and DER differential entropy regularizer.}
    \label{tab:full_ablation}
\end{table}
	
\end{document}